# CVABS: Moving Object Segmentation with Common Vector Approach for Videos

Şahin Işık [1*], Kemal Özkan [1], Ömer Nezih Gerek [2]

[1] Computer Engineering Department, Eskisehir Osmangazi University, 26480, Eskisehir, Turkey
[2] Electrical and Electronics Department, Anadolu University, 26555, Eskisehir, Turkey
*sahini@ogu.edu.tr

**Abstract:** Background modelling is a fundamental step for several real-time computer vision applications that requires security systems and monitoring. An accurate background model helps detecting activity of moving objects in the video. In this work, we have developed a new subspace based background modelling algorithm using the concept of Common Vector Approach with Gram-Schmidt orthogonalization. Once the background model that involves the common characteristic of different views corresponding to the same scene is acquired, a smart foreground detection and background updating procedure is applied based on dynamic control parameters. A variety of experiments is conducted on different problem types related to dynamic backgrounds. Several types of metrics are utilized as objective measures and the obtained visual results are judged subjectively. It was observed that the proposed method stands successfully for all problem types reported on CDNet2014 dataset by updating the background frames with a self-learning feedback mechanism.

## 1. Introduction

Separation of the foreground from background on a processed image, namely background modelling, plays an important role and positively affects performance of certain computer vison applications. It is considered as pre-process for many tasks including moving object recognition, person tracking, traffic monitoring, motion capturing, teleconference and security surveillance systems. A developed background subtraction algorithm can usually be utilized for the following tasks: monitoring the behaviours of enemy's crafts, tanks and spies with a smart surveillance system; protecting land and railways; detecting people in cameras of urban surveillance systems; and recording the behaviour of tourists, walkouts, protests, bullies and drivers by setting up a centralized spy video surveillance network in campus', street or urban areas. Concentrating an object tracking system on a developed background modelling algorithm to gather information for the prevention of crime is cheap, simple and more efficient.

Video backgrounds can be considered in two categories as static and dynamic backgrounds. While static backgrounds do not change in time, dynamic ones can be exposed by continuous, irregular or abrupt changes. Such cases usually occur due to illumination variations, shadow, motion of camera and others effects (such as surging of water, motion of tree leaves in windy weather, and vibrated scenes) [1]. Obviously, the methods optimized for static backgrounds do not perform well on dynamic backgrounds [2-5]. One solution for the dynamic background problem is to continuously update the background model according to a set of controller parameters of the dynamic. A crowded set of methods have been proposed in the dynamic background modelling literature. While some existing theories rely on subspace based background modelling, many state of the art methods consider smart feedback mechanisms to update the backgrounds.

A subspace learning approach that was based on variants and improvements of Principal Component Analysis (PCA) was investigated for background subtraction under illumination changes in [6]. An eigenspace model was formed by taking a sample of N images in the In SL-PCA method [7]. Since the obtained eigenspace model was not adequate to describe the portions of an image containing a moving object, a more robust model of the probability distribution function of the background was determined by computing the sum of eigenbasis vectors related to the static portions of the image in [7]. Later, background modelling was carried out by estimating the de-mixing matrix $\mathbf{W} = (\mathbf{W}_{i1}, \mathbf{W}_{i2})^T$ in the SL-ICA method [8]. The projection of de-mixing matrix onto the mixing matrix, $\mathbf{X}_T = (\mathbf{X}_B, \mathbf{X}_F)^T$, gives the learned background model as denoted by $\mathbf{Y} = \mathbf{W}\mathbf{X}_T$, where $\mathbf{X}_T = (\mathbf{X}_B, \mathbf{X}_F)^T$ indicates the mixture data matrix (2xK dimensions) with $\mathbf{X}_B$ referring to background frames (vector format of background images) and $\mathbf{X}_F$ referring to foreground frames (vector format of foreground images). Additionally, in SL-INMF method, an incremental algorithm (inspired from Non-negative Matrix Factorization - NMF) was proposed for modelling the dynamic backgrounds [9]. NMF was used for reducing dimension and extracting intuitive features in an efficient, robust and simple way. From a different perspective, Li et al. have developed a new subspace based background learning strategy by utilizing a high-order tensor learning algorithm, namely incremental rank-(R1,R2,R3) tensor [10]. In an extension work, namely IRTSA, the Background Model (BM) was constructed by a low-order tensor eigenspace model, where the associated sample mean and the eigenbasis were updated adaptively [11].

Unfortunately, these subspace techniques were not suitable for real time applications due to their low robustness and high computational requirements. Therefore, faster and more memory efficient methods had been sought after. As an outcome of those efforts, fine performances for foreground detection were obtained on CDnet [12] website, such as SharedModel [13], Weight Sample Background Extractor (WeSamBE) [14], Self-Balanced SENsitivity SEgmenter (SUBSENSE) [15], Pixel-based Adaptive Word Consensus

Segmenter (PAWCS) [16], Flux Tensor with Split Gaussian Model (FTSG) [17], and Multimode Background Subtraction (MBS) [18]. Among these, SharedModel [13] considers the concept of Gaussian Mixture Model (GMM) to represent each pixel with best GMM model. In SuBSENSE [15] and PAWCS [16] methods, a powerful feedback mechanism was performed to update backgrounds online for every processed frame. While SuBSENSE [15] was concerned with utilizing Local Binary Similarity Pattern (LBSP) features, PAWCS [16] deals with colour/LBSP features to generate dictionaries for sample representations and segmentation. FTSG [17] algorithm reveals the ability of Flux Tensor based motion segmentation along with predefined rules and chamfer matching based validation of foreground regions. Using a single colour model, MBS [18] algorithm investigates the capability of using two colour models (RGB and YCbCr) in order to extract foreground regions. Once two foreground masks were acquired from test image, Aggregation/Fusion of morphological processes were applied on the obtained binary foreground masks to yield a meaningful output.

This paper presents a background modelling algorithm that aims to meet expectations due to different variations occurring at vibrated backgrounds along time. To illustrate; light variation in the background is high in the morning time in relation to the sun's rays whereas this variation disappears in the evening time. Therefore, the performance of computer vision based applications including motion detection, object tracking, and surveillance is substantially affected by such conditions. Typically, accuracies of indoor surveillance systems present different trends depending upon the light conditions. To alleviate the problems caused by illumination, a new subspace-based background subtraction model has been developed.

To this aim, an unorthodox, smart and subspace-based algorithm for background modelling is presented here, by utilizing the concept of Common Vector Approach (CVA) obtained with the Gram-Schmidt Orthogonalization procedure [19-22]. The key principle of the study is to consider background modelling procedure as a spatio-temporal classification problem, i.e., the first class is foreground and the other is background. From this viewpoint, the background of an image set can be acquired by computing the associated common vector of the frames. Starting with this idea, we employed the CVA on a matrix in which each column vector refers to different frames. Once a unique common vector that encapsulates the general characteristic of the related class, called background model, is subtracted from the discriminative common vector of the processed frame, a feedback mechanism is carried out to reduce the negative effects of dynamic scenes and illumination on performance for foreground object detection. Since the concept of a CVA based background subtraction is performed for a moving object segmentation, we have called the algorithm as CVABS. The detail work is explained in the following related chapters.

The rest of paper is organized as follows. Section 2 presents an introduction to the CVA method and describes the procedures for how the method was adopted for background modelling. Section 3 focuses on the similarities and differences between the common vector and average vector. Section 4 summarizes the process of foreground detection and background modelling with dynamic controller parameters. Then, in Section 5, the performance of the proposed method on several datasets is discussed and compared with state of the art methods based on subjective and objective evaluation. Finally, conclusions are presented in the last section.

## 2. Proposed Method
### 2.1. Principle of CVA

CVA [19] is a subspace based recognition method that gives satisfactory results in a variety of applications and classification tasks including face [20], spam e-mail [21], edge detection [22]. The derivation of CVA is inherited from the idea behind the Karhunen Loewe Transform (KLT). While, in KLT, the projection is taken onto the eigenvector corresponding to the largest eigenvalues, in the CVA method, this procedure is carried out in the opposite direction by projecting the data onto the eigenvector associated with the smallest eigenvalues. While the eigenvectors with largest eigenvalues correspond to the most descriptive axes, the eigenvectors with smallest (or zero) eigenvectors provide the axes that share the common characteristics of the class (with no variations). The advantage of CVA over other subspace based classification methods is the derivation of a solution for a, so called, insufficient data case, which occurs when the dimension of the feature vector is greater than the number of data samples [23]. This, for example, cannot provide a solution case with Fisher Linear Discriminant Analysis (FLDA), where computing the inverse of within-class correlation matrix would not be possible. When the background separation problem is considered, it can be noticed that the number of pixels in each frame is naturally way larger than the number of frames in a sequence, rendering this case a solid example of insufficient data case.

The insufficient case in CVA is handled via Gram-Schmidt orthogonalization. Let us suppose that we are given $k$ samples corresponding to the $i$-th class (different sequential views of the same scene, $\{\mathbf{a}_j^i\}$ $j = 1, 2, ..., k$). The class index could be any integer, so let's drop the superscript $i$ without any loss of generality. It is now possible to represent each $\mathbf{a}_j$ vector as the sum of $\mathbf{a}_j = \mathbf{a}_{i,com} + \mathbf{a}_{i,diff}$. Here, a common vector ($\mathbf{a}_{i,com}$) is what is left when the difference vectors are removed from class members and is invariant throughout the class, whereas $\mathbf{a}_{i,diff}$ is called the remaining vector, which represents the particular residual trend of this particular sample. There are two cases in CVA where the number of vectors is either sufficient or insufficient. In this study, we have focused on the insufficient data case since the frames are handled as a vector format.

Let us now suppose that the training set has k samples $(\mathbf{a}_1, \mathbf{a}_2, ..., \mathbf{a}_k)$ corresponding to the $i$-th class in $\mathbb{R}^k$. Also, each sample has a dimension of $h \times w$. To find a common vector for any class, we should construct a matrix from these samples. Hence, the column matrix with the $(h \times w) \cdot k$ dimension is obtained from the given $k$ samples. Our aim is to project the column matrix onto the 1-D space (vector), by preserving the global (hence common) information. To understand this, the algorithm described below should be followed based on the rules given in the related work on CVA [19].

- First, a random vector, i.e., $\mathbf{a_1}$ is taken as a reference, then the difference vectors belonging to the processed data is obtained by:

$$\mathbf{d_{j-1}} = \mathbf{a_j} - \mathbf{a_1} \quad j = 2, 3, ..., k \quad (1)$$

- Once *(k-1)* difference vectors are obtained, the difference subspace (DS) for *i-th* class can be calculated by gathering the difference vectors.

$$\mathbf{DS_i} = \{\mathbf{d_j}, \mathbf{d_{j+1}}, ..., \mathbf{d_{k-1}}\} \quad j = 1, 2, ..., k-1 \quad (2)$$

- In the next stage, the Gram-Schmidt orthogonalization procedure is applied to $\mathbf{DS_i}$ to obtain the orthonormal basis, $(\mathbf{z_1}, \mathbf{z_2}, ..., \mathbf{z_{(k-1)}})$ which spans the difference subspace and orthogonalizes the difference vectors of the *i-th* class. The obtained orthogonal vectors are divided by their corresponding Frobenius Norm's to make them normalized, producing an orthonormal basis.

- In $\mathbb{R}^{k-1}$, the orthonormal basis $(\mathbf{z_1}, \mathbf{z_2}, ..., \mathbf{z_{(k-1)}})$ and orthogonal vectors $(\mathbf{v_1}, \mathbf{v_2}, ..., \mathbf{v_{(k-1)}})$ of the plane with *k* dimensions are computed with the following formulas.

$$\mathbf{v_1} = \mathbf{d_1} \quad and \quad \mathbf{z_1} = \frac{\mathbf{v_1}}{\|\mathbf{v_1}\|}$$

$$\mathbf{v_2} = \mathbf{d_2} - \langle \mathbf{d_2}, \mathbf{z_2} \rangle \mathbf{z_2} \quad and \quad \mathbf{z_2} = \frac{\mathbf{v_2}}{\|\mathbf{v_2}\|} \quad (3)$$

$$\mathbf{v_3} = \mathbf{d_3} - \langle \mathbf{d_3}, \mathbf{z_2} \rangle \mathbf{z_2} - \langle \mathbf{d_3}, \mathbf{z_1} \rangle \mathbf{z_1} \quad and \quad \mathbf{z_3} = \frac{\mathbf{v_3}}{\|\mathbf{v_3}\|}$$

...

$$\mathbf{v_j} = \mathbf{d_j} - \sum_{j=1}^{j-1} \langle \mathbf{d_j}, \mathbf{z_{j-1}} \rangle \mathbf{z_{j-1}} \quad and \quad \mathbf{z_j} = \frac{\mathbf{v_j}}{\|\mathbf{v_j}\|} \quad j = 2, ..., k$$

$(\mathbf{z_1}, \mathbf{z_2}, ..., \mathbf{z_{(k-1)}})$ refers to an orthonormal basis for and $\mathbb{R}^{k-1}$ for a given plane, respectively. Here, $\langle .,. \rangle$ implies the inner product of the given vectors and $\|.\|$ denotes the norm of the vectors. Since, in general, these vectors may have a dimensionality of more than 2, Frobenius norm is used here. Naturally, $(\mathbf{v_1}, \mathbf{v_2}, ..., \mathbf{v_{(k-1)}})$ refers to orthogonal vectors for a given plane and for $\mathbb{R}^{k-1}$, respectively.

- Once the orthonormal basis are computed, the difference vectors $\mathbf{a_{i,diff}}$ can be obtained by the projection of any sample $\mathbf{a_j}$ from the *i-th* class on the difference subspace of a class which is spanned by a orthonormal basis $(\mathbf{z_1}, \mathbf{z_2}, ..., \mathbf{z_{(k-1)}})$:

$$\mathbf{a_{i,diff}} = \langle \mathbf{a_j}, \mathbf{z_1} \rangle \mathbf{z_1} + \langle \mathbf{a_j}, \mathbf{z_2} \rangle \mathbf{z_2} + \cdots + \langle \mathbf{a_j}, \mathbf{z_{k-1}} \rangle \mathbf{z_{k-1}} \quad (4)$$

- Finally, as shown in Eq. (5), subtracting the $\mathbf{a_{i,diff}}$ from any vector $\mathbf{a_j}$, gives a common vector of the i-th class. Practically, any sample among the $(\mathbf{a_1}, \mathbf{a_2}, ..., \mathbf{a_k})$ can be used as a reference. By considering the given form of $\mathbf{a_j} = \mathbf{a_{i,com}} + \mathbf{a_{i,diff}}$, the common vector can be formulated as;

$$\mathbf{a_{i,com}} = \mathbf{a_j} - \mathbf{a_{i,diff}} \quad (5)$$

producing $\mathbf{a_{i,com}}$ which refers to a common matrix of the *i-th* class. Thus, a class with several samples can be represented by a unique subspace called a *common vector*.

To summarize, the projection of vectors established from each sample of a class onto an orthonormal basis gives the difference vectors. If the difference vectors are subtracted from the reference vector, the common vector of the processed class is acquired.

### 2.2. Application to Background Modelling

Our main objective is to combine the common characteristic stated in different sequential video frames with a single view that reserves the rich information about the background. Using the above CVA algorithm to represent the different views with a common one can be considered as a similar procedure to the work of Oliver et. al, called SL-PCA method [7]. When cross-referenced to the study, the eigenspace model generated from the PCA decomposition is utilized by considering the fact that moving objects do not appear in static regions which are the contributions of moving objects to the eigenspace model and are very small and can even be negligible. With this concept, the difference between the mean background ($\mathbf{u}$) and the column representation of each input image ($\mathbf{I_t}$) was projected onto the $h$ dimensional eigen-background subspace, ($\mathbf{\Phi}_h$), which consists of the eigenvectors associated to the largest eigenvalues of a column representation of the $k$ frames, denoted with $\mathbf{B_t}$. In the following step, the ($\mathbf{I_t}$) has been reconstructed to represent the background model ($\mathbf{I_t^{'}}$) as shown in Eq. (7).

$$\mathbf{B_t} = \mathbf{\Phi}_h (\mathbf{I_t} - \mathbf{u}) \quad (6)$$

$$\mathbf{I_t^{'}} = \mathbf{\Phi}_h^T \mathbf{B_t} + \mathbf{u} \quad (7)$$

Finally, those foreground pixels related to a moving object are detected by considering the distance between the input ($\mathbf{I_t}$) and the reconstructed background ($\mathbf{I_t'}$) frames regarding the predefined threshold *T* as denoted with the rule below;

$$\mathbf{F_t(i,j)} = \begin{cases} 1 & if \ dist(\mathbf{I_t}(i,j)), \mathbf{I_t'}(i,j)) > Threshold \\ 0 & else \end{cases} \quad (8)$$

The procedure stated for the SL-PCA is the inspiration for our study in background subtraction. In that aspect, our work considers the problem from the reverse perspective; we are considering the background as the eigenspace with lower variations. Specifically, the common vector of a column representation of the k frames can be obtained either by using the eigenspace model that consists of eigenvectors corresponding to the smallest eigenvalues, or by obtaining the orthonormal vectors of the processed data with the Gram-Schmidt orthogonalization procedure in the case of insufficient data. Since obtaining the eigenvector for a large dimension of data requires an enormous computer memory, we have concentrated on the Gram-Schmidt orthogonalization procedure instead of deriving the orthonormal vectors.

Although using the CVA algorithm for background modelling gives good results in case of low correlated data with rank greater than 2, it has been observed that the CVA concept collapses in case of high correlated data ranked near 2. For example, it is favourable to obtain a nice common vector (background model) in the case of 'backdoor' video sequence, but it is obvious that the non-meaningful common vectors derived from the '*copyMachine*', '*office*' and '*library*' video sets (in which the sequential frames are too similar to each other) cause problems. With these problematic videos, the value range of the common vector becomes different from 0-255, causing segmentation problems since the difference between the test and the common vector does not reveal the accurate foreground regions (see Fig. 1).

By taking these adverse effects of pure CVA into account, we have put together a new CVA methodology to obtain an accurate distance map between the test and background frames. For this purpose, as an extension to CVA, the Discriminative Common Vector Approach (DCVA) option has been considered in distance computation stage. The DCVA method is naturally inherited from CVA. First, the common vector of the background frames is obtained with the above explained CVA method, then the discriminative common vector is obtained by taking the projection of the test vector onto the orthonormal vectors generated from the Gram-Schmidt procedure. During classification, the L1 norm was observed to give better results for the foreground motion, as can be seen in Fig. 1. Detailed information about the foreground segmentation process is provided in chapter, *Foreground Extraction*.

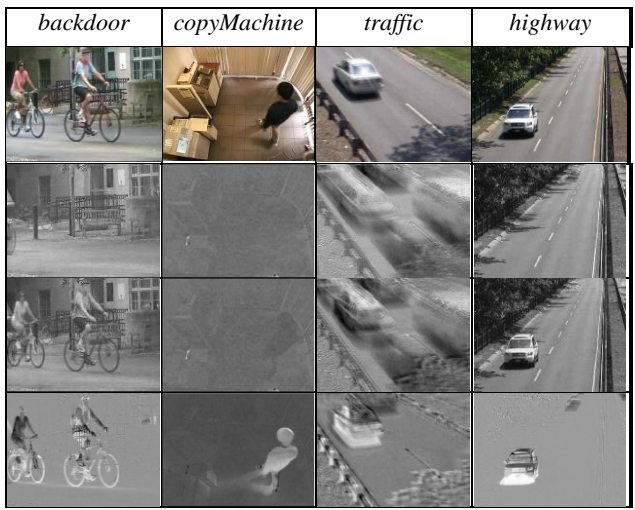

**Fig. 1.** The visual demonstration of common frame of backgrounds, discriminative common frame and distance map between them.

Fig. 1 shows examples of visual results from CVABS in terms of background modelling. The first row presents the test frames. The second row of Fig. 1 exhibits the common frame (background model) derived from the first 35 background frames. Then, the third row shows the visualization of the discriminative common frame examples related to the test image to be processed for moving object detection. The last row displays the distance between the discriminative common frame and background model (common frame), which clearly exposes the motions (hence foregrounds) in the test frame. One can observe that using the CVABS promises highly relatable results in terms of highlighting the foreground regions. With respect to the idea of the CVABS, it is expected that the common and unvarying characteristic of static regions would combine within the background model (common frame) while details such as unstable regions including illuminations, reflections and waving trees would transfer to the difference frame.

The discriminative common vector related to the test frame $a_t$ can be achieved through the following steps.

- Once First, the difference vector $a_{t,diff}$ is obtained after the projection of the test vector onto the orthonormal basis $(z_1, z_2, ..., z_{(k-1)})$.

$$a_{t,diff} = \langle a_t, z_1 \rangle + \langle a_t, z_2 \rangle + \cdots + \langle a_t, z_{k-1} \rangle \quad (8)$$

- Then, as shown in Eq. (10), subtracting the $a_{t,diff}$ from the test vector ($a_t$), gives a discriminative common vector associated to $a_t$.

$$a_{t,com} = a_t - a_{t,diff} \quad (10)$$

## 3. Common Vector Versus Average Vector

As aforementioned, the common vector is obtained by taking the difference between the average vector and sum of the projection of average vector onto the orthonormal basis. Consequently, it is clear that the common vector is not the average vector. To explain the contribution of common vector against the plain average, a mathematical proof is illustrated and the difference in performance is compared with a simple numerical example and a visual demonstration.

Let's assume that we are given *m* vectors and $a_i$, $a_{com}$ and $z_i$ refers to a training vector, the common vector and orthonormal basis returned from Gram-Schmidt Orthogonalization, respectively. Hence, all the vectors in the training set can be written with following forms:

$$a_1 = a_{com} + <a_1, z_1> z_1 + <a_1, z_2> z_2 + ... + <a_1, z_{m-1}> z_{m-1}$$

$$a_2 = a_{com} + <a_2, z_1> z_1 + <a_2, z_2> z_2 + ... + <a_2, z_{m-1}> z_{m-1}$$
$$\ldots \quad (11)$$
$$a_m = a_{com} + <a_m, z_1> z_1 + <a_m, z_2> z_2 + ... + <a_m, z_{m-1}> z_{m-1}$$

If we summarize both sides of the above equation side by side, we can obtain:

$$\sum_{i=1}^{m} a_i = m\,a_{com} + < \sum_{i=1}^{m} a_i, z_1 > z_1 + < \sum_{i=1}^{m} a_i, z_2 > z_2 +$$

$$... + < \sum_{i=1}^{m} a_i, z_{m-1} > z_{m-1}$$

$$\frac{1}{m}\sum_{i=1}^{m} a_i = a_{com} + < \frac{1}{m}\sum_{i=1}^{m} a_i, z_1 > z_1 + < \frac{1}{m}\sum_{i=1}^{m} a_i, z_2 > z_2 +$$

$$... + < \frac{1}{m}\sum_{i=1}^{m} a_i, z_{m-1} > z_{m-1}$$

and $\quad (12)$

$$\mathbf{a}_{com} = \mathbf{a}_{ave} - <\mathbf{a}_{ave}, \mathbf{z}_1> \mathbf{z}_1 - <\mathbf{a}_{ave}, \mathbf{z}_2> \mathbf{z}_2 - \ldots - <\mathbf{a}_{ave}, \mathbf{z}_{m-1}> \mathbf{z}_{m-1}$$

As we can see, the common vector is obtained by subtracting the average vector from its projection onto the entire orthonormal basis. To analyse the behaviours of an average and a common vector on a toy example of three simple training vectors, let's suppose that the 3 vectors are $\mathbf{a}_1 = \begin{bmatrix} 1 & 1 & 1 \end{bmatrix}^T$, $\mathbf{a}_2 = \begin{bmatrix} 1 & 1 & -1 \end{bmatrix}^T$ and $\mathbf{a}_3 = \begin{bmatrix} 1 & 5 & 5 \end{bmatrix}^T$. Let's also assume that the closest vectors ($\mathbf{a}_1$ and $\mathbf{a}_2$) refer to the background and the distinct vector ($\mathbf{a}_3$) refers to the foreground. The common vector of these three vectors can be obtained through the difference vectors ($\mathbf{b}_1$ and $\mathbf{b}_2$) by taking $\mathbf{a}_1$ as a reference:

$$\mathbf{b}_1 = \begin{bmatrix} 0 & 0 & -2 \end{bmatrix}^T \text{ and } \mathbf{b}_2 = \begin{bmatrix} 0 & 4 & 4 \end{bmatrix}^T \quad (13)$$

Then we proceed with the Gram-Schmidt Orthogonalization and the common vector of the training set is acquired by the following rules:

$$\mathbf{d}_1 = \mathbf{b}_1, \quad \mathbf{z}_1 = \frac{\mathbf{b}_1}{\|\mathbf{b}_1\|} = \begin{bmatrix} 0 & 0 & -1 \end{bmatrix}^T \quad (14)$$

$$\mathbf{d}_2 = \mathbf{b}_2 - <\mathbf{b}_2, \mathbf{z}_1> \mathbf{z}_1 = \begin{bmatrix} 0 & 4 & 0 \end{bmatrix}^T$$

$$\mathbf{z}_2 = \frac{\mathbf{d}_2}{\|\mathbf{d}_2\|} = \begin{bmatrix} 0 & 1 & 0 \end{bmatrix}^T \quad (15)$$

$(\mathbf{z}_1, \mathbf{z}_2, \ldots, \mathbf{z}_{(k-1)})$ shows the orthonormal basis and $(\mathbf{b}_1, \mathbf{b}_2, \ldots, \mathbf{b}_{(k-1)})$ indicates the orthogonal vectors. Hence, the summation of the projections of $\mathbf{a}_1$ onto the orthonormal basis of the difference subspace $\mathbf{B}$, which is denoted with $\mathbf{a}_{sum}$, can be obtained as follows:

$$\mathbf{a}_{sum} = <\mathbf{a}_1, \mathbf{z}_1> \mathbf{z}_1 + <\mathbf{a}_1, \mathbf{z}_2> \mathbf{z}_2 = \begin{bmatrix} 0 & 1 & 1 \end{bmatrix}^T \quad (16)$$

Finally, the common vector can be obtained by subtracting $\mathbf{a}_{sum}$ from either the reference vector ($\mathbf{a}_1$) or the average vector. By using the reference vector, the common vector can be obtained as in Eq. (17).

$$\mathbf{a}_{com} = \mathbf{a}_1 - \mathbf{a}_{sum} = \begin{bmatrix} 1 & 0 & 0 \end{bmatrix}^T \quad (17)$$

In the same vector list, by using the average vector, the common vector could also be obtained Let us define the average vector as:

$$\mathbf{a}_{ave} = \sum_{i=1}^{m} \mathbf{a}_i = \begin{bmatrix} 1 & 2.33 & 1.67 \end{bmatrix}^T \quad (18)$$

then project onto the orthonormal basis:

$$\mathbf{a}_{ave,sum} = <\mathbf{a}_{ave}, \mathbf{z}_1> \mathbf{z}_1 + <\mathbf{a}_{ave}, \mathbf{z}_2> \mathbf{z}_2 + \ldots + <\mathbf{a}_{ave}, \mathbf{z}_{m-1}> \mathbf{z}_{m-1} \quad (19)$$

$$= \begin{bmatrix} 0 & 2.33 & 1.67 \end{bmatrix}^T$$

Hence, the common vector would be:

$$\mathbf{a}_{com} = \mathbf{a}_{ave} - \mathbf{a}_{ave,sum} = \begin{bmatrix} 1 & 0 & 0 \end{bmatrix}^T \quad (20)$$

which is precisely the same vector found from Eq. 17. Yet, if the difference magnitudes between the training vectors, and average and common vector are computed, we will observe the different distances.

$$F_1^{ave} = \|\mathbf{a}_1 - \mathbf{a}_{ave}\|^2 = 1.4907, \quad F_1^{com} = \|\mathbf{a}_1 - \mathbf{a}_{com}\|^2 = 1.4142$$

$$F_2^{ave} = \|\mathbf{a}_2 - \mathbf{a}_{ave}\|^2 = 2.9814 \quad F_2^{com} = \|\mathbf{a}_2 - \mathbf{a}_{com}\|^2 = 1.4142 \quad (21)$$

$$F_3^{ave} = \|\mathbf{a}_3 - \mathbf{a}_{ave}\|^2 = 4.2687, \quad F_3^{com} = \|\mathbf{a}_3 - \mathbf{a}_{com}\|^2 = 7.0711$$

These different distance results also indicate that the common vector and average vector perform completely differently for classification performances. When the common vectors are used, results similar to the ones in Fig. 1 are obtained, whereas simple averaging provides nothing other than simple blurring, yielding unsuccessful background-foreground separation. To compare the performances of an average and a common vector, the obtained F values are taken as a reference. For the best performance, we would like to expect that the obtained distance results are close to one another for the background model (i.e., $\mathbf{a}_1$ and $\mathbf{a}_2$), and the distance for the foreground vector ($\mathbf{a}_3$) should be as high as possible. Clearly, $F_1^{com}$ and $F_2^{com}$ are the same with $F_3^{com}$ being significantly a larger, whilst $F_1^{ave}$ and $F_2^{ave}$ are not much alike with $F_3^{ave}$ not being much different. The toy example also motivates that the CVA can separate the foreground from the background with a high-performance rate when compared with an average vector.

## 4. CVA Implementation for Foreground Segmentation and Background Updating

### 4.1. Foreground Extraction

As outlined in the above sections, the common frame associated with background frames is obtained for background modelling. The main objective is to derive a robust distance between the test frames and the background frames with respect to a CVABS based motion segmentation procedure. During initialization, the first N frames are considered as a background list called $B(x) = \{B_1(x), \ldots, B_k(x), \ldots, B_N(x)\}$, where $x$ corresponds to a pixel location and Bi are the frames. Conventionally [24], using the first $N = 35$ frames for initialization of the background bank is suitable, when considering the speed and performance. Eq. (22) represents the foreground detection module utilized to determine the foreground map between the test and the background frames.

$$F_t(x) = \begin{cases} 1 & if \#\left\{\sum_{i=1}^{N} dist\left(I_t(x), B_i(x)\right) \geq R(x)\right\} > \#min \\ 0 & otherwise \end{cases} \quad (22)$$

where the '#' operator is used as a counter of the pixels. Therefore, $\#min$ is a decision threshold to assign the label of a pixel as a foreground or a background, which is classically set as $\#min = N - 1$. The label of a pixel is considered as a foreground (1) if it is marked as 1 in all binary output maps ($F_1(x), F_2(x), \ldots, F_N(x)$); otherwise it is assigned as a background (0). With this strict decision threshold process, it is ensured that the algorithm becomes more robust to the noisy pixels. In the same equation, $R(x)$ is a grey level threshold to generate binary output maps.

It is widely known that the performance of all background/foreground separation algorithms depends on the utilized distance metric. In this study, we applied a hybrid distance metric consisting of three metric approaches. The first distance metric is $\ell_1$ norm distance, which is commonly used in the literature for background subtraction. However, the $\ell_1$ distance is sensitive to sudden illumination changes, which affects a large number of pixels at once. Therefore, as a second distance metric, the Gradient information is taken into account to bottle with the sudden illumination changes and shadows. The third (last) distance is based on the common vector approach concept. The $\ell_1$ distance can be notated as:

$$dist_{\ell_1}\left(I_t(x), B_i(x)\right) = |I_t(x) - B_i(x)| \quad (23)$$

which is applied by comparing the grey values of each pixel. The Gradient information (the second metric) uses the following directional gradients:

$$I_{gx'}(x) = D_{11}(x) \cdot I_{gx}(x) + D_{12}(x) \cdot I_{gy}(x)$$
$$I_{gy'}(x) = D_{12}(x) \cdot I_{gx}(x) + D_{22}(x) \cdot I_{gy}(x)$$
$$I_{gt,t}^m(x) = \sqrt{I_{gx'}(x)^2 + I_{gy'}(x)^2}$$
$$\bar{B}_{gx'}(x) = D_{11}(x) \cdot \bar{B}_{gx}(x) + D_{12}(x) \cdot \bar{B}_{gy}(x)$$
$$\bar{B}_{gy'}(x) = D_{12}(x) \cdot \bar{B}_{gx}(x) + D_{22}(x) \cdot \bar{B}_{gy}(x) \quad (24)$$
$$\bar{B}_{gt,t}^m(x) = \sqrt{\bar{B}_{gx'}(x)^2 + \bar{B}_{gy'}(x)^2}$$
$$dist_{Gmag}\left(I_t(x), B_i(x)\right) = |I_{gt,t}^m(x) - \bar{B}_{gt,t}^m(x)| \quad (25)$$

These gradient values are obtained from the test and the background frames. The utilized gradient distance metric is calculated with an edge suppression based gradient transformation approach [25], shown in Eq. (24) and Eq. (25). Firstly, the horizontal gradient map, $I_{gx}(x)$, and vertical gradient map $I_{gy}(x)$ of the test frame ($I_t$) and the mean background frame ($\bar{B}_t$) are computed with the Sobel operator. Later, the cross-diffusion tensor terms, called $D_{11}$, $D_{22}$ and $D_{12}$, are determined with respect to the rules given in the referred study. After applying the cross-diffusion tensor terms to $I_t$ and $\bar{B}_t$, then the gradient transformed versions, $I_{gt,t}^m(x)$ and $\bar{B}_{gt,t}^m(x)$, are acquired to compute gradient distance. The absolute distance between these two new robust gradient maps, $I_{gt,t}^m(x)$ and $\bar{B}_{gt,t}^m(x)$, gives accurate and noise free foreground localization. It is known that the illuminations and shadows generally distribute in a homogeneous way. Therefore, using the gradient information between test and background frames improves the performance by diminishing the sudden illumination effect. Owing to this, the homogeneous regions are suppressed by taking the derivation procedure. In this study, a new gradient distance metric is computed to reduce the ghost problem caused by sudden changes and intermittent object motion problems.

The third and main distance metric is expressed with a CVABS based distance computation:

$$dist_{cva}\left(I_t(x), B_i(x)\right) = |I_{d,com}(x) - B_{com}(x)| \quad (26)$$

As explained before, first, the common frames related to the background frames are determined with respect to the CVA method. Then, the discriminative common frame of the test frame is computed by taking projection of the test frame onto the orthonormal vectors associated with the background list. As shown in Eq. (26), the absolute difference between the common frame and the discriminative common frame gives us a novel distance metric for foreground detection.

Finally, Eq. (27) demonstrates the final distance metric as a combination of the three distance metrics:

$$dist\left(I_t(x), B_i(x)\right) = dist_{\ell_1}\left(I_t(x), B_i(x)\right)$$
$$+ dist_{Gmag}\left(I_t(x), B_i(x)\right) + dist_{com}\left(I_t(x), B_i(x)\right) \quad (27)$$
$$= |I_t(x) - B_i(x)| + |I_{gt,t}^m(x) - \bar{B}_{gt,t}^m(x)| + |I_{d,com}(x) - B_{com}(x)|$$

As a hybrid distance metric; the pure grey level distance, gradient distance and CVA distance are combined to obtain a robust and weighted distance term. The condition of $dist_{\ell_1}\left(I_t(x), B_i(x)\right) > 1$ is carried out in the final distance metric to refuse the noisy regions caused by dynamic scenes. By using this final distance metric, pixel values related to foreground regions take higher values and the value of noisy and unwanted pixels becomes lower. Besides, the gradient transformation enables us to wipe out the ghosts and illuminations. Once the distance is calculated with this way, the segmentation process is applied to the final distance map to determine the label of each pixel. The CVA distance together with the traditional and gradient distance is observed to generate plausible results in challenging videos.

### 4.2. Updating Background Frames

Once the foreground is detected, the background frames should be updated in a smart way, which is robust against unstable dynamics acting on the frames. Technically, the selected methodology for the background frames' updating process should avoid a leakage of foreground regions into the background frames (or vice versa). In this study, we have adopted a widely utilized and proven way, which relies on the statistical histogram information about foreground pixels through time. The histogram of foreground regions, which reflects to the *learning rate* (*T*), is calculated as an accumulator whose value changes based on the dynamic's controller parameters. After segmentation of the final distance between the test and background frame, the class belonging probability rate is computed online as $p \approx 1/T(x)$ for each pixel. The benefits of updating pixels with respect to a probability rate is that the static foreground regions would not enter into the background for a long enough time. The videos containing static foreground regions such as *copyMachine*, *library* and *office* are well-known representations for static videos. In such videos, a person waits for minutes to complete a task related to the real world, i.e., reading a book in a *library* or waiting to use a busy *copyMachine*. If necessary precaution is not considered, such static objects would leak to the background class after some frames. Conversely, dynamic regions such as waving trees and sea waves would easily leak to the background class with respect to the utilized probability rate, $p$. These are avoided as follows.

Classically, there are $N$ background frames to be updated during the process. However, we update only a randomly selected frame among the $N$ background frames after segmentation of each test frame at time $t$. As shown in Eq. (28), the randomly selected frame at time $t$, called $B_t$, is updated by considering the probability rate, $p$, related to each pixel location ($x$) of test frame and background frame:

$$B_t(x) = (1-p) \cdot B_t(x) + p \cdot I_t(x) \quad (28)$$

where the $I_t(x)$ refers to the processed pixel of the incoming test frame at time $t$. $B_t(x)$ denotes the background's pixel to be updated at time $t$. This background update mechanism depending on the foreground histogram helps incorrect classifications, however further learning rate adjustments and adaptive threshold selections are necessary, as will be explained *in Sec. 4.3*.

### 4.3. Monitoring Learning Rate (T(x)) and Threshold (R(x)) with Dynamic's Controllers

As introduced in the foreground detection and background updating processes, there are two key parameters utilized in the CVABS algorithm. The first parameter is called the learning rate, $T(x)$, that specifies how much rate of a pixel is permitted to enter the background. The other important functional parameter is given as the decision threshold, $R(x)$, which arranges the threshold for each pixel when encountering a sudden illumination change. By considering recommendations suggested by [15, 26, 27] for monitoring the $T(x)$ and $R(x)$, we have utilized two dynamic's controller parameters ($d_{min,t}(x)$ and $v(x)$) to achieve the correct segmentation of frames.

Fig. 2 presents the benefits of the utilized self-learning feedback mechanism in terms of monitoring the utilized internal parameters. To show the activation of the smart feedback mechanism, some examples from *fountain01*, *cubicle*, *winterDriveway* are chosen as representative videos for challenging backgrounds including water waves, sudden illuminations and intermittent object motion, respectively.

From Fig. 2, we can observe that there is a great difference between the performance of the segmentation in the presence/absence of post processing and feedback mechanism. Clearly, one can note that water waves and sunbeams are marked as foreground in the absence of a feedback mechanism (column no 3 and 4). However, as can be seen from the fifth and sixth columns, the unwanted pixels can be removed in the foreground map by introducing a self-learning feedback mechanism with dynamic controller parameters. At the last step, some standard morphological operations are utilized as post processes to further remove speckle noises and obtain well segmented objects.

*4.3.1 Monitoring the Decision Threshold (R(x)):* Traditionally, some predefined decision threshold, $R(x)$, or adaptive thresholding techniques including Otsu's thresholding were considered for classifying the pixel as foreground (1) or background (0). However, the performance of fixed thresholding methods suffers from sudden distortions in videos. For example, we always witness sudden illumination changes in real-life videos when the sun appears on a sunny day or when clouds abruptly block sunlight. This situation is even more encountered in cases of turning-on the light, car or traffic lights, blinking lamps, etc. Such examples negatively affect performance by increasing the false positive rates. To overcome such problems, the past information of the distances map between test and background frames are taken into account in some state-of-the-art methods like PBAS [26], PWACS [16] and SUBSENSE [15]. In these methods, an adaptive and self-regulated decision threshold, $R(x)$, is built through time to regulate the decision threshold with respect to sudden changes. In this study, we also employed this effective method to update the decision threshold, $R(x)$. Specifically, the algorithm works as follows.

Assuming that $d(x) = \{d_1(x), \ldots, d_k(x), \ldots, d_n(x)\}$, is the list of minimal distance to generate binary output maps over the time domain. A distance for each pixel is computed as $d_t(x) = min\{dist(I_t(x), B_i(x))\}$ at time $t$. The average value of the minimal decision list encapsulates sudden intensity changes and is called the dynamic's control parameter, computed as $d_{min,t}(x) = mean\{d_t(x), \ldots, d_{t-n+1}(x)\}$ at current time, $t$. With experimental evaluation, one can observe that the value of $d_{min,t}(x)$ would not change through the time domain while $d_{min,t}(x)$ associated to unstable sudden illumination regions and waving trees increases or decreases based on the distance metric. Usage of this feedback mechanism has numerous advantages in terms of reaching superior performance results.

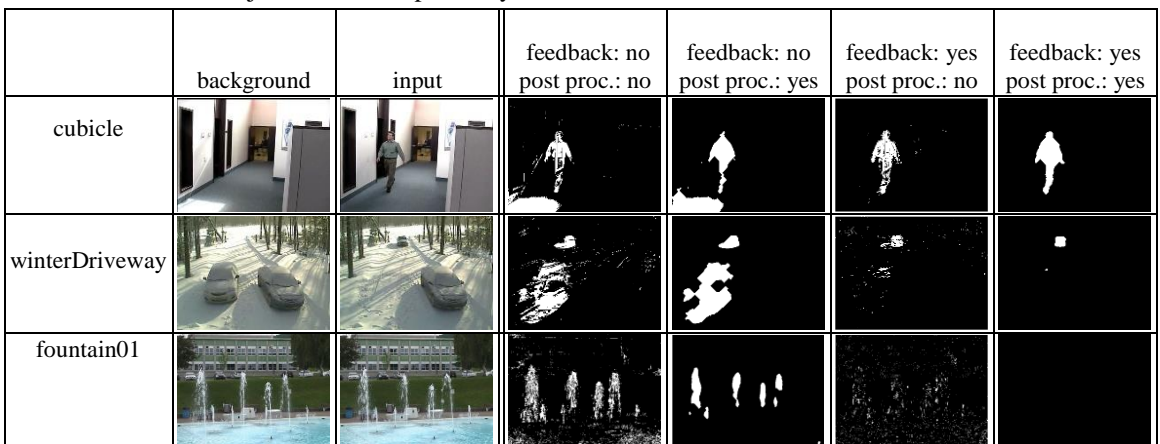

**Fig. 2**. The presentation of advantages of using dynamic control parameters and post processing.

Starting from the above calculation of $d_{min,t}(x)$, we monitor the decision threshold, $R(x)$ according to Eq. (29), where the value of the decision threshold, $R(x)$, for each pixel in the test frame is steadily arranged by the dynamic's controller parameter, $d_{min,t}(x)$.

$$R(x) = \begin{cases} R(x) + R_{inc/dec} \cdot R(x) & \text{if } R(x) < (d_{min,t}(x) \cdot R_{scale}) \\ R(x) - R_{inc/dec} \cdot R(x) & \text{otherwise} \end{cases} \quad (29)$$

Since the value of $R(x)$ changes steadily, we have to determine bounds for $R(x)$, that is $R_{lower} \leq R(x) \leq R_{upper}$. The lowest value of $R(x)$, $R_{lower}$, is constant and predefined (a common initial value is 35), and the upper value of $R(x)$ is taken as $R_{upper} = \infty$. In Eq. (29), the $R_{inc/dec}$ indicates a steering coefficient, set to 0.01 for our experiments. The steering decision threshold is a scaled version of $d_{min,t}(x)$. The scale (i.e., $R_{scale}$) of this threshold is selected from a set of three values, 0.1, 1 and 2, according to the complexity of the processed video. Experimentally, we have observed that the value of $R_{scale}$ should be taken as 0.1 for simple and static videos. For complex videos, the value of $R_{scale}$ should be increased to 1 or even to 2 for even more complicated scenes such as *backdoor, cubicle, fall* or *fountain01* videos.

### 4.3.2 Monitoring the Learning Rate (R(x)):

As mentioned before, some objects can stay steadily at a constant position for a long period. This may cause the system to mistakenly tag the object as background. Usually these regions are marked as foreground at some past frame time, then they leak into the background category, even with systems having slow updating parameter. Such pixels generate false alarms and are also the reasons for the, so called, ghosting phenomenon, where pixels are marked as foreground but not associated to any objects in the test image, or are tagged as background with an abrupt movement. Examples of ghost regions include shadows, corrupted background's pixels and intermittent object motions. A typical example of a ghost problem is the stopped car in *winterDriveway* as shown in Fig. 2. Ghost regions increase false positive rates and decrease the overall performance of the moving object segmentation algorithm. To avoid ghosts, the regions of steadily marked foregrounds should be updated with a relatively small probability rate, $p$.

To further reduce ghost region effect, a common parameter, called the learning rate ($T$), is introduced to properly update the background frames. In this study, the $T$ parameter was updated based on $v$ and $\hat{d}_{min,t}$ parameters, which are also the dynamic's controllers. Parameter $\hat{d}_{min,t}$ refers to a normalized version of $d_{min,t}$ in an interval of [0-1]:

$$T(x) = \begin{cases} T(x) + \dfrac{1}{v(x) \cdot \hat{d}_{min,t}(x) + 1} & \text{if } F_t(x) = 1 \\ T(x) - \dfrac{v(x) + 0.1}{\hat{d}_{min,t}(x) + 1} & \text{if } F_t(x) = 0 \end{cases} \quad (30)$$

In Eq. (30), the $F_t$ indicates that the segmentation map corresponding to the test image at time t. Similar to the threshold, the learning rate, $T$ is also bounded to $T_{lower} \leq T(x) \leq T_{upper}$. The lower bound is $T_{lower} = 2$ while the upper bound is taken as $T_{upper} = \infty$. From Eq. (30), we can see that, for each pixel, the value of $T(x)$ is updated with $\hat{d}_{min}(x)$ and $v(x)$ parameters, according to the class belongings ($F_t(x) = 0$) or ($F_t(x) = 1$). Here, the $v(x)$ parameter is a positive-valued accumulator that holds histogram information for blinking foreground pixels. It implicitly works as follows; for dynamic regions including waving trees and sea waves, the value of $v(x)$ is increased by +1, for static regions, the value of $v(x)$ is decremented by -0.1 and converges to 0, as can be seen in Eq. (31).

$$v(x) = \begin{cases} v(x) + 1 & \text{if } X_t(x) = 1 \\ v(x) - 0.1 & \text{if } X_t(x) = 0 \end{cases} \quad (31)$$

where $X_t(x)$ corresponds to the binary map obtained by the XOR operation between the consecutive foreground maps, current foreground map ($F_t(x)$) and previous foreground map ($F_{t-1}(x)$). Also, the regions intersected with $F_t(x)$ and $F_{t-1}(x)$ are set to 0 to avoid the leaking of the foreground's borders into the background frame.

Using the calculated $v(x)$, we can see that the pixels associated to a high value of $v(x)$ would reduce the learning rate for background pixels because $\left( (v(x) + 0.1) / (\hat{d}_{min}(x) + 1) \right)$ is subtracted from $T(x)$ to immediately update the background pixels related to blinking pixels. If the value of $v(x)$ is large in case of a foreground pixel, then the learning rate will slowly increase because $v(x)$ appears in the denominator of the additional term. With the above use of the dynamic's controller parameters, the encountered noisy regions including waving trees, water waves in fountains and camera motions would be collected in background frames and marked as background for better performance.

### 4.4. Handling PTZ Motions

When comparing with other categories, one may see that the PTZ category is particularly complicated and involves elaborate methods to obtain well-segmented foreground regions. The reasons can be attributed to unsteady background with pan, tilt, and zoom operations, yet alone shaky and jittered frames. To come up with a solution for this category, we have introduced a scene change detection algorithm [28]. The scene change case is triggered from fast statistical measurements including Mean Absolute Edge Difference (MAED), Mean Absolute Frame Difference (MAFD) and Absolute Difference Frame Variance (ADFV) over time. To catch an instant scene change, the conditions given in Eq. (32) have to be supplied as touched in the study of [24]. Once a scene change is detected, the background / foreground formation process is reset.

$$MAED > 0.1, \quad MAFD > 30, \quad ADFV > 2 \quad (32)$$

## 5. Experimental Study

### 5.1. Dataset

The experimental works were conducted on the Change Detection (CDnet 2014) Dataset [29] in order to analyse the performance of the proposed methods. The reason for this particular choice that this dataset can attribute a large variety of background types related to real world events including *badWeather, baseline, cameraJitter, dynamicBackground, intermittentObjectMotion, lowFramerate, night Videos, PTZ, shadow, thermal* and *turbulence*. For each class, we have used the first $N$ (set to 35) samples for the initialization of the background model bank.

### 5.2. Performance Evaluation on CDnet 2014 Dataset

Several objective and subjective metrics have been applied in order to analyse the performance of the proposed method. As originally noted in the study of CDnet, possible statistical metrics to measure the performance of the background subtraction techniques are computed based on the number of True Positives (TP), number of True Negatives (TN), number of False Negatives (FN) and number of False Positives (FP) by using the predefined ground truth images for each video type given in CDnet 2014. Eventually, a combined metric of Matthew Correlation Coefficient (MCC) is obtained as

$$MCC = \frac{(TP \cdot TN) - (FP \cdot FN)}{\sqrt{(TP+FP)\cdot(TP+FN)\cdot(TN+FP)\cdot(TN+FN)}}. \quad (33)$$

An alternative metric is called the F-score, which depends on two other metrics, namely the precision and recall, according to Eq. 34:

$$F - score = \frac{2 \cdot precision \cdot recall}{precision + recall} \quad (34)$$

In this study, we have considered the MCC as [15] and F-score [12] metrics to compare the performance of CVABS with other state of art methods in an objective way. The F-score is in the interval [0-100] in terms of percentage [30] and MCC is a balanced metric that reveals the correlation between the two binary samples and its value between -1 and 1, with 1 corresponding to the best performance.

To give a general insight into the performance of the CVABS method, the visual outputs of several state of art methods are demonstrated and compared in Fig. 3. As a common evaluation method, the judgements are expressed in terms of visual inspection of foreground masks, that is the obtained segmentation results are compared with ground truth (GT; second row of images) in Fig. 3. By visual inspection, one can observe that the CVABS method gives well segmented and satisfactory results to cope with the illuminations and dynamic changes. The CVABS advocates the reduction of unstable light effects especially for shadow videos like cubicle. Again, from the fall and turbulence0 videos, which include intense effects of dynamic scenes, we can observe that the utilized dynamic controller parameters perform very well in terms of accurate update of the backgrounds and reduction of false alarms. To reveal the strength of CVABS to sustain the performance in case of turbulence degradations, let us compare the foreground segmentation results of each method over the *turbulence0* video. We can instantly see that the CVABS gives clean results for the given sample of the *turbulence0* video, whereas the SharedModel, PAWCS and MBS methods show certain performance impairments. We also see that WeSamBE, SuBSENSE and FTSG generate similar outputs with the ground truth for this video due to their insufficient update strategies for this type of difficulty. In another case (a night video, called *streetCornerAtNight*), we can observe that most all methods (except MBS) achieve rather nice segmentation results. However, when the *winterDriveway* video is considered, we can observe that five out of seven methods (which are SharedModel, WeSamBE, PAWCS and SuBSENSE and MBS) give erroneous results, with CVABS being in the list of accurate methods.

This is basically because CVABS gradually updates background frames with utilized feedback parameters to regulate changes in frames. Thus, ghosts are alleviated based on the utilized self-regulation and self-learning procedure in terms of the background frames updating. A strong argument of the proposed method is that, CVABS always remain the class of "fair performing" foreground/background classifiers in a wide variety of cases. Therefore, it is argued that the proposed CVABS method is subjectively a top tier, and it outperforms other methods on the average.

As a presentation of objective scores, we present Fig. 4, which demonstrates the MCC and F-score for CVABS and other the top ranked methods listed in CDnet. By examining the MCC coefficients in Fig. 4, we can highlight the superior performance of CVABS with a 77.01% F-score and 77.54% MCC value when compared with others. An interesting point is that, that degradation in the F-score also results in low MCC values, with the exception of PAWCS and FTSG. The closest competitor to CVABS seems to be the SharedModel, but the clear difference of the performance of CVABS is attributed to the fact that CVABS simply does not fail in any of the difficult cases, whereas other successful methods all have at least one weak point (that varies according to the difficulty). This phenomenon reflects to the results of both MCC and F-score values. The individual F-scores of each method (over the CDnet 2014 dataset) are presented in Table 1.

By examining the results from Table 1, one can see that the CVABS method gives certain superior results in of *badWeather, lowFramerate, nightVideos* and *thermal* categories.

| | *badWeather* | *dynamic Background* | *intermittent ObjectMotion* | *nightVideos* | *shadow* | *thermal* | *turbulence* |
|---|---|---|---|---|---|---|---|
| | *blizzard* #3109 | *fall* #1889 | *winterDriveway* #2025 | *streetCorner AtNight* #902 | *cubicle* #2788 | *library* #3608 | *turbulence0* #2854 |
| *Test* | | | | | | | |
| *GT* | | | | | | | |
| *CVABS* | | | | | | | |
| *SharedModel* | | | | | | | |
| *WeSamBE* | | | | | | | |
| *PAWCS* | | | | | | | |
| *SuBSENSE* | | | | | | | |
| *MBS* | | | | | | | |
| *FTSG* | | | | | | | |

**Fig. 3**. Visual performance demonstration on the CDnet dataset.

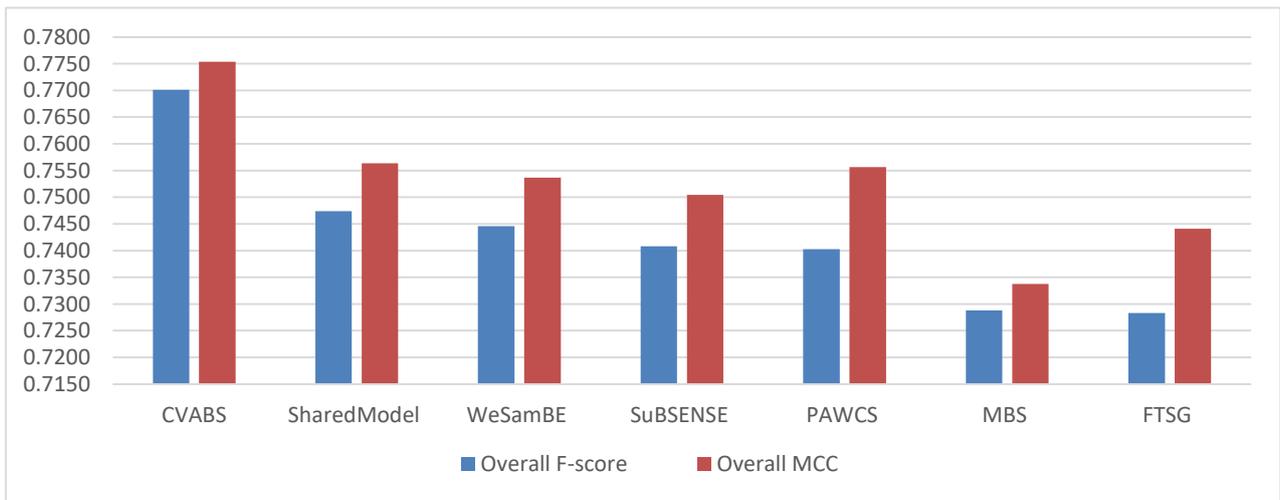

**Fig. 4**. MCC and F-score results for top ranked methods given in CDnet and CVABS.

Table 1. Performance (F-score) evaluation in more details for each category of aforementioned methods.

| | F-measure Scores of each Category (%) | | | | | | | | | | |
|---|---|---|---|---|---|---|---|---|---|---|---|
| Method | PTZ | BW | BA | CJ | DB | IOM | LF | NV | Sh | Th | Tu |
| **CVABS** | 0.4699 | **0.8570** | 0.9147 | 0.7837 | 0.8618 | 0.6586 | **0.7856** | **0.6418** | 0.8361 | **0.8567** | 0.7665 |
| **SharedModel** | 0.3860 | 0.8480 | **0.9522** | 0.8141 | 0.8222 | 0.6727 | 0.7286 | 0.5419 | 0.8455 | 0.8319 | 0.7339 |
| **WeSamBE** | 0.3844 | 0.8608 | 0.9413 | 0.7976 | 0.7440 | 0.7392 | 0.6602 | 0.5929 | **0.8999** | 0.7962 | 0.7737 |
| **SuBSENSE** | 0.3476 | 0.8619 | 0.9503 | 0.8152 | 0.8177 | 0.6569 | 0.6445 | 0.5599 | 0.8646 | 0.8171 | **0.7792** |
| **PAWCS** | 0.4615 | 0.8152 | 0.9397 | 0.8137 | **0.8938** | 0.7764 | 0.6588 | 0.4152 | 0.8710 | 0.8324 | 0.6450 |
| **MBS** | **0.5520** | 0.7980 | 0.9287 | **0.8367** | 0.7915 | 0.7568 | 0.6350 | 0.5158 | 0.8262 | 0.8194 | 0.5858 |
| **FTSG** | 0.3241 | 0.8228 | 0.9330 | 0.7513 | 0.8792 | **0.7891** | 0.6259 | 0.5130 | 0.8535 | 0.7768 | 0.7127 |

In the other test categories, other methods occasionally outperform CVABS. For example, in the PTZ category, the highest F-score value is obtained with the MBS method. OR, the FTSG method gives the best results in intermittent object motions. However, the critical point to mention is that, even for cases where CVABS may not be at the top, it always appears in the top tier list. Other methods fail to have this property. For example, although MBS is at the top for the *PTZ* category, that method is at the bottom of the list for *badWeather (BW), dynamicBackground (DB)* and *lowFramerate (LF)* categories. This failure does not happen in the proposed CVABS method.

## 6. Conclusion

In this study, a foreground/background extraction method is proposed as a combination of a novel Common Vector Approach (CVA) and further state-of-the-art post processing techniques. The CVA method is observed to provide a very reasonable background candidate that needs to be fine-tuned by carefully adjusting the class membership thresholds. The post-processing step of the proposed method considers a pixel adaptive segmentation and updating process with the aid of an internal feedback mechanism. Objective and subjective evaluations are performed over the CDnet 2014 data set with a large variety of video types. The results of recent and proven methods are compared to the performances of the proposed method (CVABS) according to the ground truth data. Experimental results show that the ability of the proposed method is sufficient to cope with dynamic variation and illumination changes in terms of background modelling. The proposed method is superior to other methods in terms of average F-score and MCC values for the CDnet dataset. A particular point we emphasize that, the proposed CVABS method never fails to the bottom of the list for any of the challenges (unlike other methods, which all have their strong and weak points). Note that, the all of presented F-scores are related to online testing on non-public samples of changedetection.net 2014 dataset. Even under the simulation medium of MATLAB, the run time of the proposed algorithm was found to be about 0.1 second per each frame with size 240x320 pixels. The execution time includes processes of reading and writing of images, distance computation, segmentation, post processing, update of internal parameters and update of backgrounds, working on an ordinary PC (Intel core i7-6700HQ with 2.60 GHz CPU and 8 GB memory). It is argued that the method can be optimized further with a faster compiled implementation with, say, OpenCV and C++. Such fast implementations and testing of several more subspace methods for background subtraction remain as future studies for this work. Other details about CVABS are available in a GitHub repository: *https://github.com/isahhin/cvabs*.